\documentclass[times,twocolumn,final,authoryear]{elsarticle}

\usepackage{prletters}
\usepackage{framed,multirow}

\usepackage{amssymb}
\usepackage{latexsym}
\usepackage{graphicx}
\usepackage{amsmath}
\usepackage{array}
\usepackage{subfig}
\DeclareMathOperator*{\argmin}{arg\,min}
\graphicspath{{./images/}}

\usepackage{url}
\usepackage{xcolor}
\definecolor{newcolor}{rgb}{.8,.349,.1}

\journal{Pattern Recognition Letters}

\begin{document}

\ifpreprint
  \setcounter{page}{1}
\else
  \setcounter{page}{1}
\fi

\begin{frontmatter}

\title{Residual Codean Autoencoder for Facial Attribute Analysis}

\author[]{Akshay  \snm{Sethi}} 
\author[]{Maneet \snm{ Singh}}
\author[]{Richa \snm{ Singh}}
\author[]{Mayank \snm{ Vatsa}\corref{cor1}}
\cortext[cor1]{Corresponding author: 
  Tel.: +91-9654653404;  
  fax:  +91-11-26907410;}
\ead{mayank@iiitd.ac.in}

\address{IIIT-Delhi, New Delhi, India}

\received{29 March 2017}

\begin{abstract}

Facial attributes can provide rich ancillary information which can be utilized for different applications such as targeted marketing, human computer interaction, and law enforcement. This research focuses on facial attribute prediction using a novel deep learning formulation, termed as R-Codean autoencoder. The paper first presents Cosine similarity based loss function in an autoencoder which is then incorporated into the Euclidean distance based autoencoder to formulate R-Codean. The proposed loss function thus aims to incorporate both magnitude and direction of image vectors during feature learning. Further, inspired by the utility of shortcut connections in deep models to facilitate learning of optimal parameters, without incurring the problem of vanishing gradient, the proposed formulation is extended to incorporate shortcut connections in the architecture. The proposed R-Codean autoencoder is utilized in facial attribute prediction framework which incorporates patch-based weighting mechanism for assigning higher weights to relevant patches for each attribute. The experimental results on publicly available CelebA and LFWA datasets demonstrate the efficacy of the proposed approach in addressing this challenging problem.


\end{abstract}

\begin{keyword}
\KWD Attribute Prediction \sep Cosine Similarity \sep Residual Learning \sep Deep Learning
\end{keyword}
 \end{frontmatter}


\section{Introduction}
\label{intro}

\textit{What Does Your Face Shape Say About You? \footnote{http://tinyurl.com/k5t6rh7}}

\textit{Is Your Personality Written All Over Your Face? \footnote{http://tinyurl.com/lpcz5jb}}

\begin{figure}
\begin{center}
  \includegraphics[width=\linewidth]{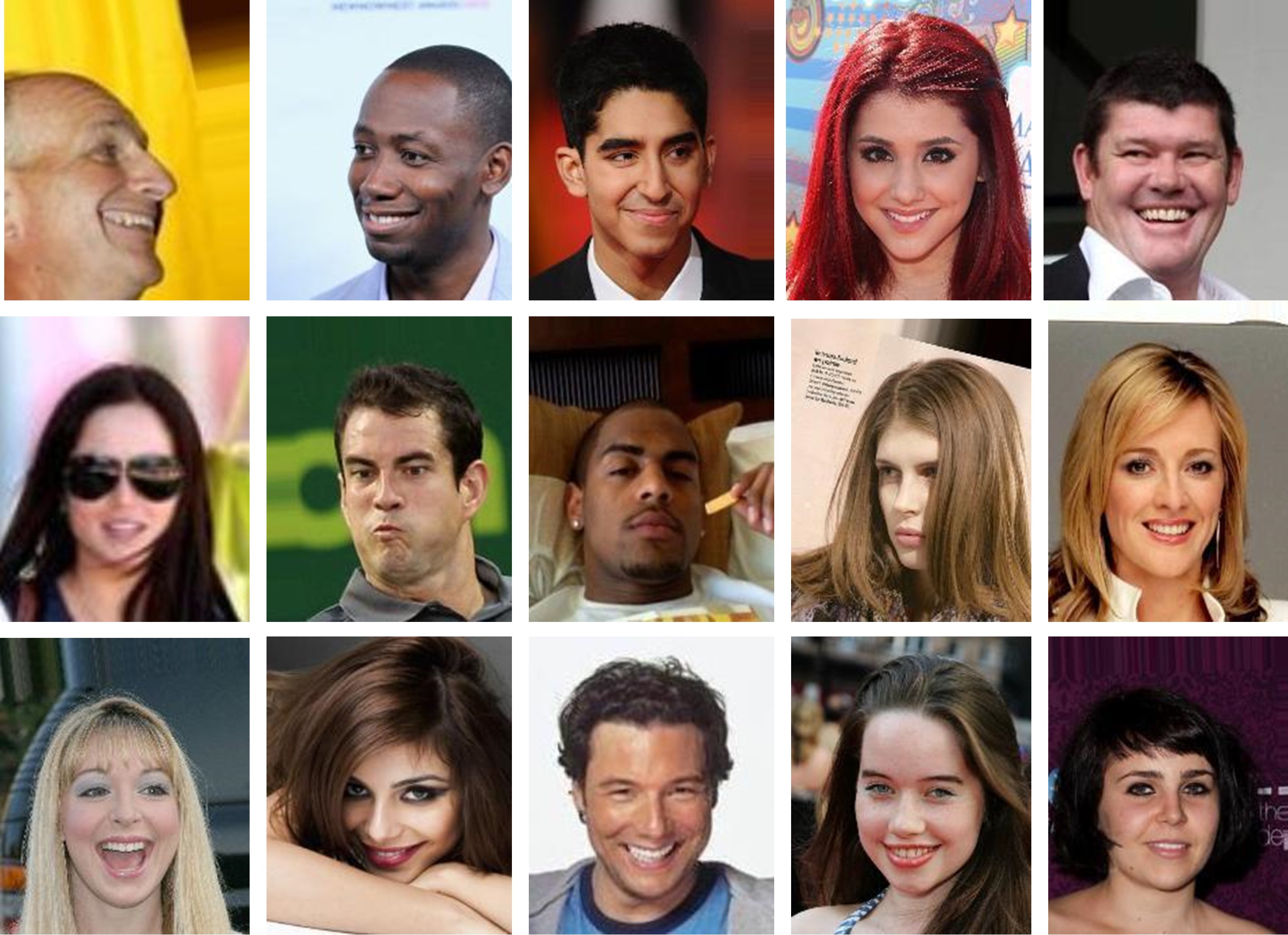}
  \caption{Sample images from the CelebA dataset \citep{liu2015deep} demonstrating intra-class variations for different attributes. The first row corresponds to the \textit{smile}, second to \textit{young}, and third to \textit{attractive}.}
  \label{fig:celeb}
  \end{center}
\end{figure}

\renewcommand{\arraystretch}{1.2}
\begin{table*} [h]
\caption{Literature review of techniques used in facial attribute prediction.}
\vspace{-5pt}
\begin{center}
 \begin{tabular}{|m{3.5cm}|m{9cm}|m{1.5cm}|m{2.5cm}|} 
 \hline
Authors &  Algorithm & Datasets Used & Classification Accuracy (Avg.) \\  
 \hline\hline
\cite{kumar2009} & Hand crafted features and RBF SVMs as classifiers  & LFW & 83.62\% \\ 
 \hline
 \cite{chung2012deep} & Supervised DBN for attribute classification & LFW & 86.00\% \\ 
 \hline
 \cite{berg2013poof} & Local part based features termed as POOF and linear SVMs & LFW & 83.00\% \\ 
 \hline
 \cite{luo2013deep} & Sum Product Network for prediction of attributes  & LFW & 87.90\% \\ 
 \hline
\cite{liu2015deep}  & Image localization deep CNN, followed by another deep CNN for prediction & CelebA, LFWA & 87.00\%, 84.00\% \\
\hline
\cite{huang2016learning} & CNN based model to address the problem of class imbalance during training & CelebA & 84.00\% \\ 
\hline
  \cite{ehrlich2016facial} & Multi Task RBMs for learning a joint feature representation for attribute classification & CelebA & 87.00\% \\ 
 \hline
 \cite{wang2016walk} & Siamese Network minimizing loss for face verification, followed by fine-tuning on CelebA dataset & CelebA, LFWA & 88.00\%, 87.00\% \\
 \hline
\cite{zhong2016face} & Features extracted from off-the shelf deep CNN models and classified using linear SVMs & CelebA, LFWA & 86.60\%, 84.70\% \\ 
 \hline
\cite{zhong2016leveraging} & Features extracted from intermediate layers of deep CNN and classified using linear SVMs & CelebA, LFWA &89.80\%,85.90\% \\  
 \hline
 \cite{rozsa2016facial} & Treats attribute classification as a regression problem and uses MSE loss to train a VGG-16 topology CNN & CelebA &90.80\% \\  
\hline
\cite{rudd2016moon} & CNN based model to perform multi task optimization with class imbalanced training data & CelebA & 90.94\% \\ 
\hline
\cite{hand2016attributes} & Multi-task deep CNN which also models attribute relationships. & CelebA, LFWA & Avg. accuracy not reported \\
\hline

\end{tabular}
\vspace{-10pt}
\end{center}
\label{tab:litRev}
\end{table*} 

These are some questions among other common questions related to facial attributes. A face image can provide multitude of information such as identity, gender, race, apparent age, and expression. While gender, race, and age estimation from face images are well explored research problems, estimating other attributes such as hair color, nose shape, eye shape, and attractiveness is also getting attention. Predicting facial attributes has several unique applications ranging from focused digital marketing to law enforcement. Facial attribute prediction has an important application in the domain of online marketing, where targeted advertisements or products can be shown to a user based on his/her physical features and appearance. The increasing usage and access of computing devices (laptops, mobile phones) and Internet facilities has also led to the availability of abundant unlabeled data, which requires tagging in order to utilize facial attributes for meaningful tasks. Further, facial attributes can be used as ancillary (or soft) information for face recognition systems, and can help in reducing the identity search space. Existing research in the literature has shown substantial improvement upon incorporating ancillary information for the task of person identification \citep{kumar2009, samangouei2015attribute,attributeSketch}. Fig. \ref{fig:celeb} presents some sample images from the Celeb Faces Attributes (CelebA) Dataset \citep{liu2015deep}. Each row contains images corresponding to a single attribute. The challenging nature of the problem in terms of high intra-class variations can be observed from these sample images.



\subsection{Related Work}

\cite{kumar2009}, in one of the initial works on facial attribute analysis, extracted hand crafted features such as edge magnitudes and gradient directions from facial regions and used the feature vector as input to a Support Vector Machine for attribute classification. Later, \cite{berg2013poof} proposed to extract local part based features termed as POOF followed by  linear SVMs for each attribute. In 2012, deep learning was explored by \cite{chung2012deep}, where a deep attribute network was built over Deep Belief Networks trained in a supervised manner. 
Later, \cite{liu2015deep} proposed a two stage training approach for addressing the task of facial attribute prediction. Given an unconstrained face image, localization was performed using a deep Convolutional Neural Network (CNN) trained in a weakly supervised manner, followed by another CNN for learning feature representation and classification. 
In 2016, \cite{wang2016walk} proposed a Siamese network which aimed at minimizing the error for the task of face verification. 
Inspired by the advancements in deep learning models, \cite{zhong2016face, zhong2016leveraging} also demonstrated that off-the-shelf features learned by CNN models pre-trained on massive facial identities, can effectively be adapted for attribute classification. The learned representations are provided as input to a binary linear SVM trained directly for all levels of representations to classify face attributes. The task of facial attribute prediction has also been posed as  regression problem \citep{rozsa2016facial}, where the authors adopt a 16 layer VGG topology while minimizing the mean squared error loss. \cite{huang2016learning} presented an approach for learning deep representation of class imbalanced data by incorporating triplet-header hinge loss in CNNs. Parallely, \cite{rudd2016moon} proposed a custom loss function for multiple attributes using a single Deep Convolutional Neural Network. The authors utilized the VGG-16 topology and obtained state-of-the-art classification results.  Recently, \cite{hand2016attributes} also proposed deep multi task CNNs for performing attribute classification. Some existing techniques for facial attribute prediction have been summarized in Table \ref{tab:litRev}.

\subsection{Research Contributions}
In this research, we propose a novel deep learning formulation for facial attribute prediction in the wild. The key contributions of this research are as follows:
\begin{itemize}
\item A novel Residual Cosine Similarity and Euclidean Distance based autoencoder, termed as \textbf{R-Codean} autoencoder is proposed. Unlike traditional autoencoders, the proposed formulation incorporates both direction and magnitude information at the time of feature extraction along with shortcut connections, thereby resulting in a residual network, 
\item The proposed R-Codean autoencoder is utilized to present a facial attribute prediction framework. The framework incorporates a patch-based weighting mechanism for providing higher weight to certain face patches for a given attribute, 
 
\item Experimental results on the Celeb Faces Attributes (CelebA) and Labeled Faces in the Wild Attributes (LFWA) datasets \citep{liu2015deep} illustrate the efficacy of the proposed model by achieving comparable results to existing deep Convolutional Neural Network (CNN) models. 
\end{itemize}
The remainder of this paper is organized as follows: the following section presents the proposed R-Codean autoencoder, followed by the proposed framework for facial attribute prediction in Section 3. Section \ref{sec:res} provides the details about the experimental protocol and evaluations, which is followed by the conclusions of this research.

\section{Proposed Residual Cosine Euclidean Autoencoder}
The proposed R-Codean autoencoder is built using a custom loss function which combines the traditionally used Euclidean  distance measure with the Cosine similarity. The proposed loss function ensures that the model minimizes the loss between the input and the reconstructed sample not only in terms of \textit{magnitude}, but also in terms of \textit{direction}. To the best of our knowledge, this is the first work which incorporates cosine loss into the feature learning process of an autoencoder. The proposed model also incorporates residual shortcut connections \citep{he2016deep} of two kinds ($symmetric$ and $cross$) into an autoencoder. 

\subsection{Euclidean Distance based Autoencoder}
\begin{figure}
\begin{center}
  \includegraphics[width=3.2in]{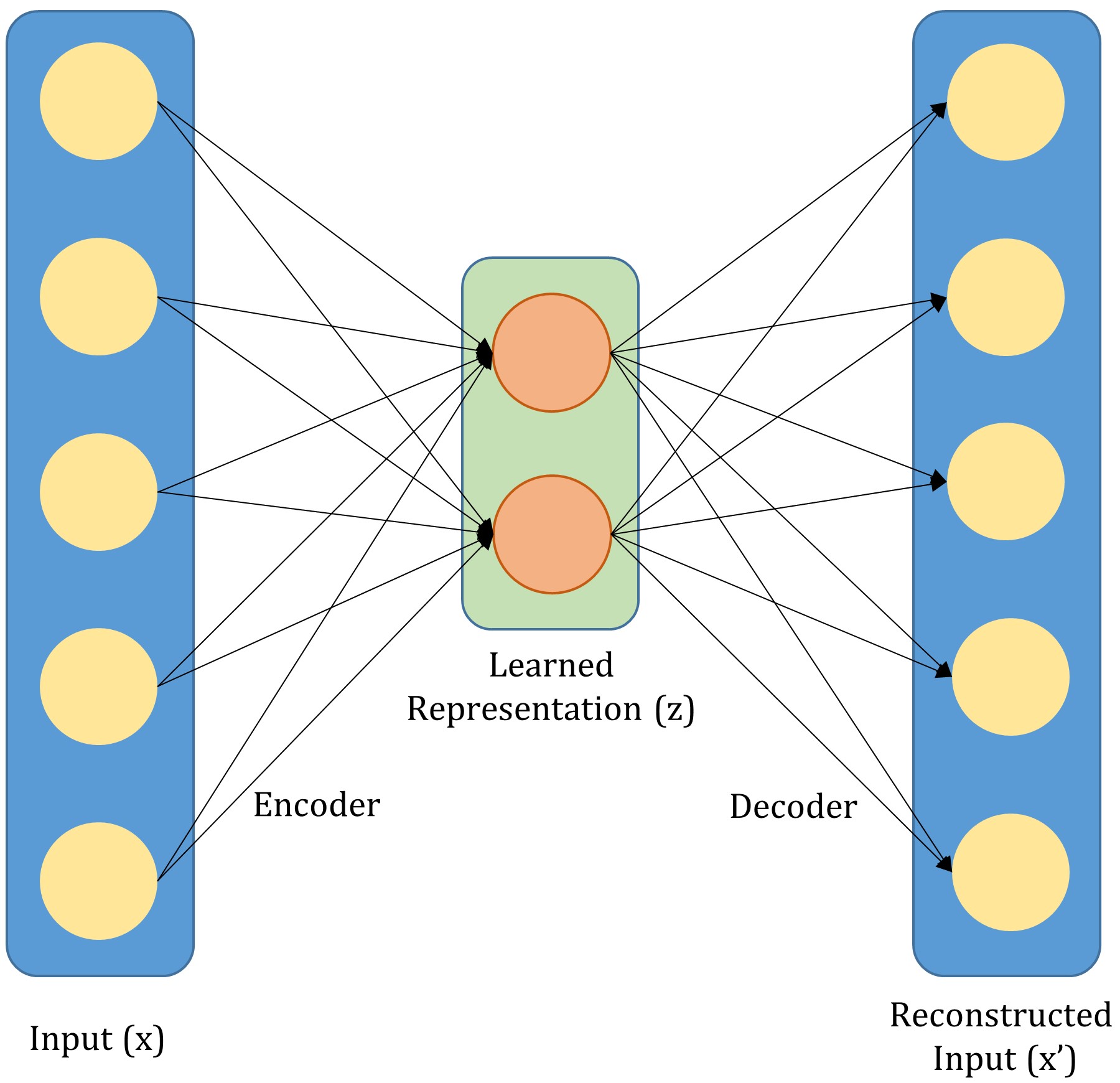}
  \caption{Diagrammatic representation of a single layer autoencoder with input $x$, learned representation $z$, and reconstructed input $x'$. }
  \label{fig:auto}
\end{center}
\end{figure}

The autoencoder model is an unsupervised deep learning architecture used for learning representations of the given data \citep{ae}. It is a special kind of neural network, where the input is also the target output. The objective of the model is to learn representations, such that the model is able to reconstruct the input from the learned representation. A single layer autoencoder consists of an encoding weight matrix ($\mathbf{W_e}$) and a decoding weight matrix ($\mathbf{W_d}$). For an input sample $x$, the loss function of an autoencoder is thus formulated as:
\begin{equation}
\argmin_{\mathbf{W_d, W_e}}⁡ \left \|x - \mathbf{W_d\ \phi (W_e}x) \ \right \|_{2}^{2}
\label{eq:euc} 
\end{equation}
where, $\phi$ corresponds to a non-linear activation function such as $sigmoid$ or $tanh$. Here, $\mathbf{W_e}x$ corresponds to the learned representation for the input $x$. The decoding weight matrix $\mathbf{W_d}$ can optionally be constrained by $\mathbf{W_d}=\mathbf{W_e}^{T}$, in which case the autoencoder is said to have tied weights. The above loss function corresponds to the mean squared error, or the Euclidean distance based loss for the autoencoder. Based on the above loss function, the model aims to minimize the reconstruction error (i.e. the error between the input $x$ and the reconstructed sample ($\mathbf{W_d\phi(W_e}x)$) in order to learn meaningful representations. Fig. \ref{fig:auto} presents a single layer autoencoder with $x$ as input, $z$ as the representation, and $x'$ as the reconstructed sample. In an attempt to learn richer feature representations, Stacked Autoencoders \citep{ae} are proposed, where multiple autoencoders are stacked on top of each other. The representation learned by the first autoencoder is provided as input to the second autoencoder which further learns higher level features. Due to the large number of parameters, deep autoencoders or stacked autoencoders are often trained in a greedy layer-by-layer manner, where only a single autoencoder is trained at a time \citep{greedy}.

\subsection{Cosine Similarity based Autoencoder}
Cosine similarity based minimization techniques have extensively been used in document modeling and retrieval \citep{doc1, doc2}. Document retrieval techniques aim to model the underlying distribution of keyword occurrences, as opposed to the actual count of the words. A similar analogy can be drawn for structured images such as faces, where cosine based similarity measure can be used for modeling the underlying distribution of the pixel values, as opposed to calculating the distance between their actual pixel intensities. To further explain, Fig. \ref{fig:dist} presents four face images of the same subject. We compute the corresponding Euclidean distance ($D_e$) and Cosine distance ($D_c$)\footnote{Cosine similarity is converted into a distance measure as $D_c = 1- S_c$, where $S_c$ is the Cosine similarity.} between the pixel values of image pairs. Since the values correspond to the distance scores, a smaller value represents more similar images. It can be observed that when the images are of the same intensity range and contain pose variations, Euclidean distance is able to model their similarity more effectively as compared to Cosine distance. On the other hand, when two images have variations in the intensity range, even with same pose, Euclidean distance is not able to encode the similarity. However, Cosine distance is able to model the similarity perfectly by producing a distance score of 0.00. Inspired by these observations, along with the properties of the Cosine similarity metric and its applicability in high dimensional feature space, we extend the formulation of traditional autoencoder to Cosine similarity based autoencoder. Cosine similarity based loss function helps the model to learn direction based features (built on the underlying distribution of pixel values) as opposed to the magnitude (pixel intensities). The loss function of a Cosine similarity based autoencoder is defined as:

\begin{figure}
\centering
\includegraphics[width=0.85\linewidth]{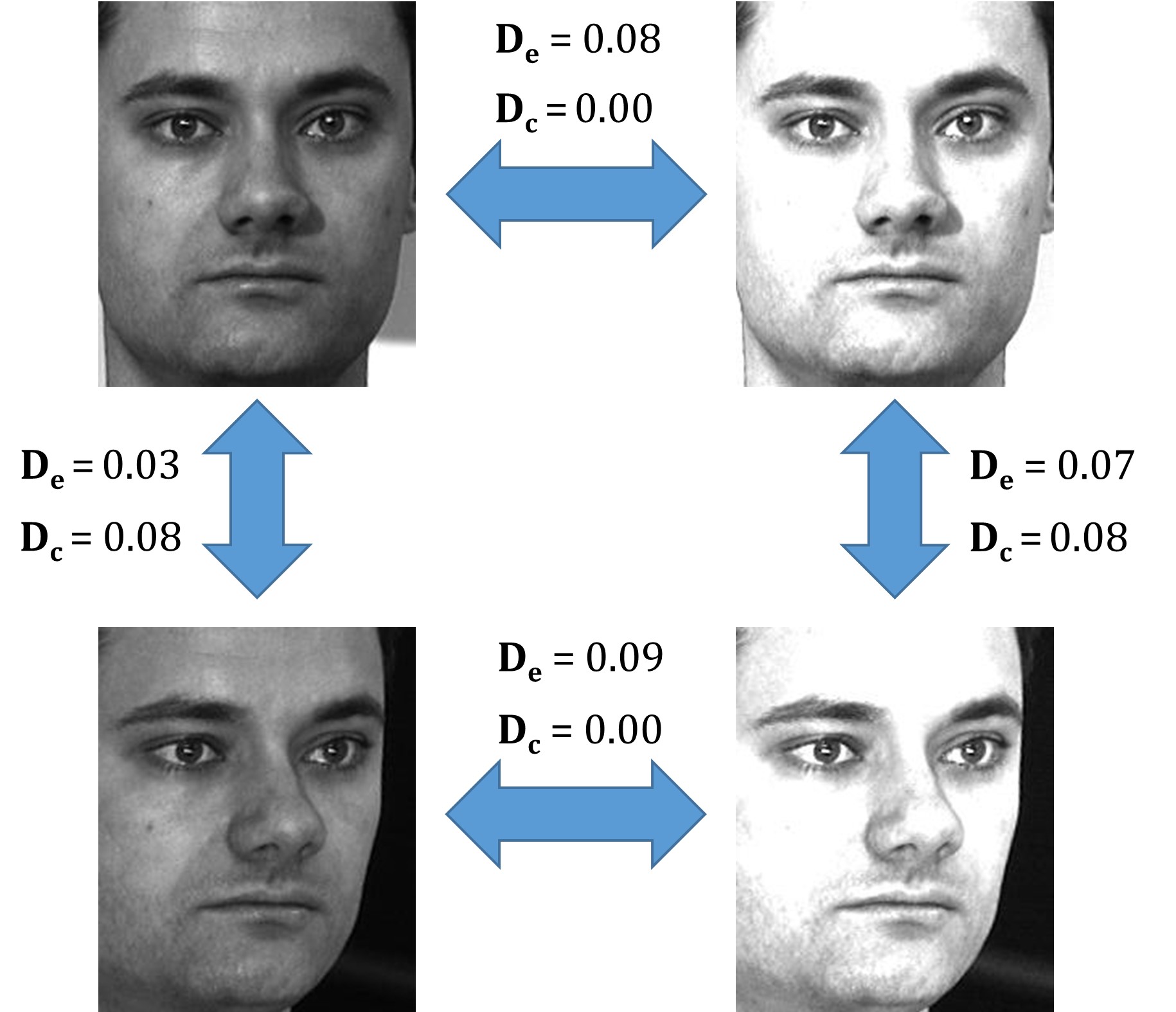}
  \caption{Sample images illustrating the effectiveness of Euclidean distance and Cosine similarity under varying conditions.  $D_e$ corresponds to the Euclidean distance between two images, while $D_c$ corresponds to the cosine distance. Cosine distance is computed as $1 - Cosine\ Similarity$. It can be observed that for samples having slight pose variations, with constant illumination, the Euclidean distance is able to model the similarity well, however, in case of illumination variations, the Cosine similarity is able to encode the similarity better. A lower distance measure corresponds to a higher similarity. }
  \label{fig:dist}
\end{figure}

\begin{equation}
\label{eq:cos}
\ell_{Cos} = -\frac{x\cdot(\mathbf{W_d\phi(W_e}x))}{ \|x\|_2^2 \times \ \|\mathbf{W_d\phi(W_e}x)\|_2^2} 
\end{equation}
where, $x$ is the input to the autoencoder and $\mathbf{W_d\phi(W_e}x)$ is the reconstructed input learned by the autoencoder. As mentioned previously, $\mathbf{W_e}$ and $\mathbf{W_d}$ represent the encoding and decoding weights of the autoencoder, respectively. It is important to note that since the Cosine metric is a $similarity$ metric, a negative sign has been incorporated in the loss function in order to decrease the overall distance, or reconstruction error of the model. The above equation minimizes the angle between the input and it's reconstruction by minimizing the cosine distance between the two. This model is especially useful for handling images with brightness or contrast variations. 

\subsection{Proposed R-Codean Autoencoder}
In order to learn feature representations based on both direction and magnitude, we combine the Euclidean distance based loss function with the Cosine similarity based loss function, and a Cosine Euclidean (Codean) Autoencoder is proposed. The loss function of the model is formulated as:
\begin{equation}
\label{eq:propLoss}
\ell_{Codean} = \alpha \times \ell_{Euc} + \beta \times \ell_{Cos} + \ \lambda R
\end{equation} 
here, the first term corresponds to the Euclidean loss (Eq. \ref{eq:euc}), the second term refers to the Cosine loss (Eq. \ref{eq:cos}), and the third term corresponds to a regularization constraint. $\alpha$, $\beta$, and $\lambda$ are the regularization parameters controlling the weight given to each individual term. The above equation helps the autoencoder learn a representation which minimizes both, the magnitude by the Euclidean loss and the direction by the Cosine similarity between the input and reconstructed sample. As explained previously (from Fig. \ref{fig:dist}), the Euclidean distance ensures that slight variations in pose in the reconstructed sample are handled by the autoencoder, while the Cosine distance handles illumination variations. For a single layer model, with input $x$, Eq. \ref{eq:propLoss} with $\ell_1$-norm regularization on the encoding weight matrix, can be expanded as follows:
\begin{equation}
\begin{gathered}
\label{eq:expanded}
\ell_{Codean} = \alpha \times  \left \|x-\ (\mathbf{W_d\phi(W_e}x)) \right \|^2_2  - \beta \times \frac{x\cdot(\mathbf{W_d\phi(W_e}x))}{ \|x\|_2^2 \times \ \|\mathbf{W_d\phi(W_e}x)\|_2^2} \\
+ \lambda \times  \left \|\mathbf{W_e}\right \|_1
\end{gathered}
\end{equation} 
The $\ell_1$-norm regularization helps learn sparse feature representations for the given input, thereby retaining meaningful latent variables during the feature learning process. The above equation depicts a single layer Codean autoencoder, which can easily be extended for $k$ layers as follows:
\begin{equation}
\begin{gathered}
\label{eq:expanded}
\ell_{Codean} = \alpha \times  \left \|x - g\circ f(x) \right \|^2_2 \ - \ \beta \times \frac{x\cdot(g\circ f(x))}{ \|x\|_2^2 \times \ \|g\circ f(x)\|_2^2} \\
+ \lambda \times \sum_{i=1}^k\left \|\mathbf{W_e^i}\right \|_1 
\end{gathered}
\end{equation} 
where, $f(x)$ is the encoder function, such that $f(x) = \mathbf{W_e^k...\phi(W_e^2(\phi(W_e^1}x)))$ and $g(x)$ is the decoder function, such that $g(x) = \mathbf{W_d^1(...(W_d^{k-1}(W_d^k}x)))$. Since the above loss function contains large number of parameters, training the entire model together results in the problem of vanishing gradients. Further, inspired by the analysis that adding shortcut connections facilitates learning of deeper networks, along with resulting in the network imitating the performance of multiple shallow networks \citep{veit2016residual}, we next propose to incorporate shortcut connections in Codean to learn robust feature representations.

\subsection{Residual Learning in Codean Autoencoder: R-Codean}
\begin{figure}
\centering
\includegraphics[width=\linewidth]{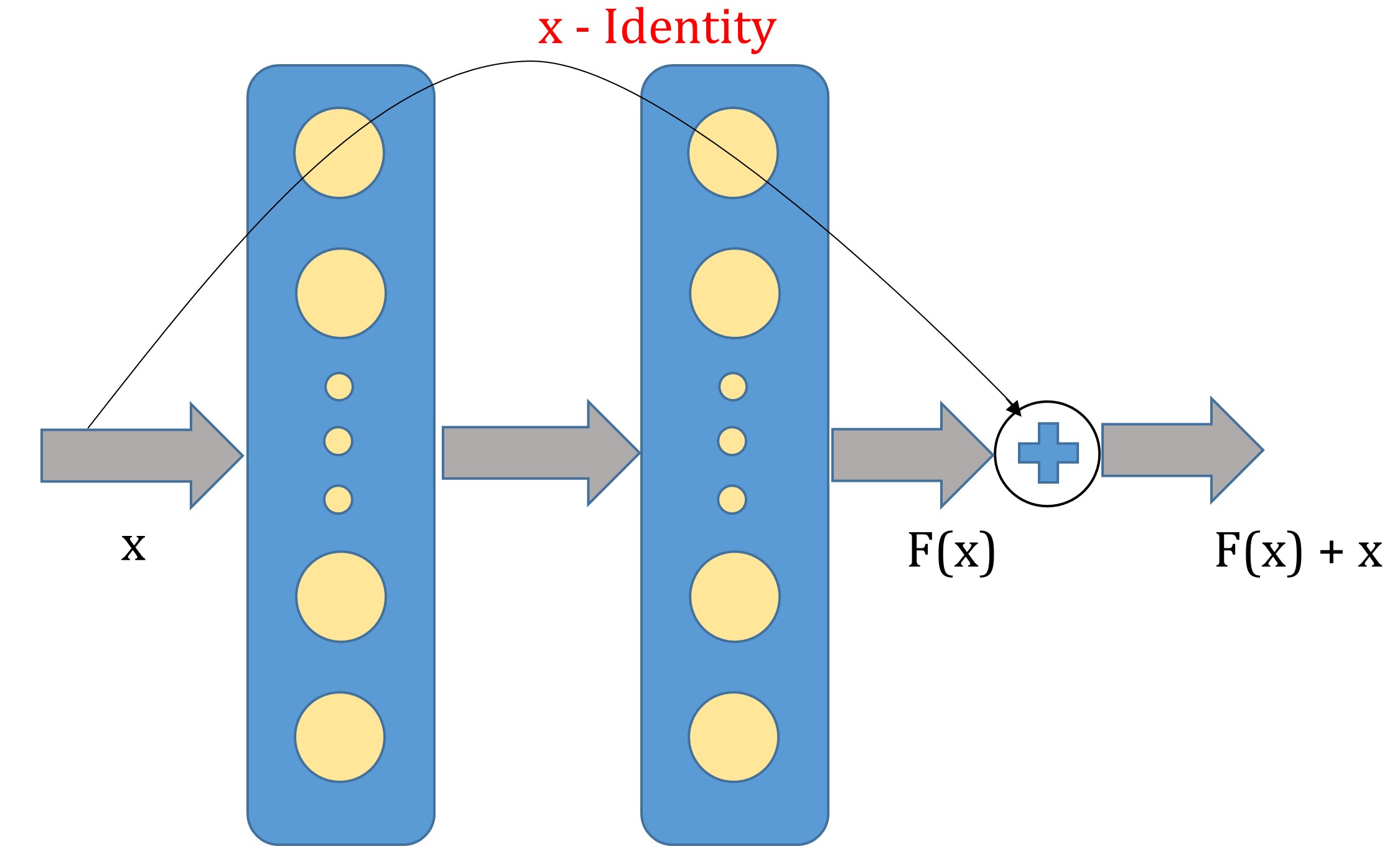}
  \caption{Shortcut connection between two layers, representing the concept of residual learning. }
  \label{fig:residual}
\end{figure}

Residual deep learning framework is introduced by He et al. \citep{he2016deep} for Convolutional Neural Networks. The authors have observed that learning deeper models resulted in a higher error rate, as compared to their shallower counterpart. Since the aim of learning deeper models is to learn higher level features with each increasing layer, shortcut (or skip) connections are introduced in an attempt to reduce the overall error and facilitate better learning. The introduction of shortcut connections in deep networks thus results in creation of residual networks. Let $H(x)$ be the mapping to be learned by a model on an input $x$. Now, instead of learning $H(x)$, the residual network is made to learn the mapping $F(x)$, where $F(x)$ is given by:
\begin{equation}
F(x) = H(x) -x
\end{equation}
The input ($x$) is then added back to $F(x)$, effectively learning $H(x)$. Residual learning as described above helps to overcome the input degradation (or vanishing gradient) problem in deep networks. In the $i^{th}$ layer of a neural network having weight matrix as $\mathbf{W}_i$, the residual learning framework can be incorporated with a building block defined as:
\begin{equation} \label{eq5}
y=F(x,\mathbf{W}_i) + x
\end{equation} 
here, $x$ and $y$ are the input and output vectors of the $i^{th}$ layer. The function $F(x, \mathbf{W}_i)$ represents the mapping to be learned. Fig. \ref{fig:residual} presents a sample residual network of two layers.

\begin{figure}
  \includegraphics[width=\linewidth]{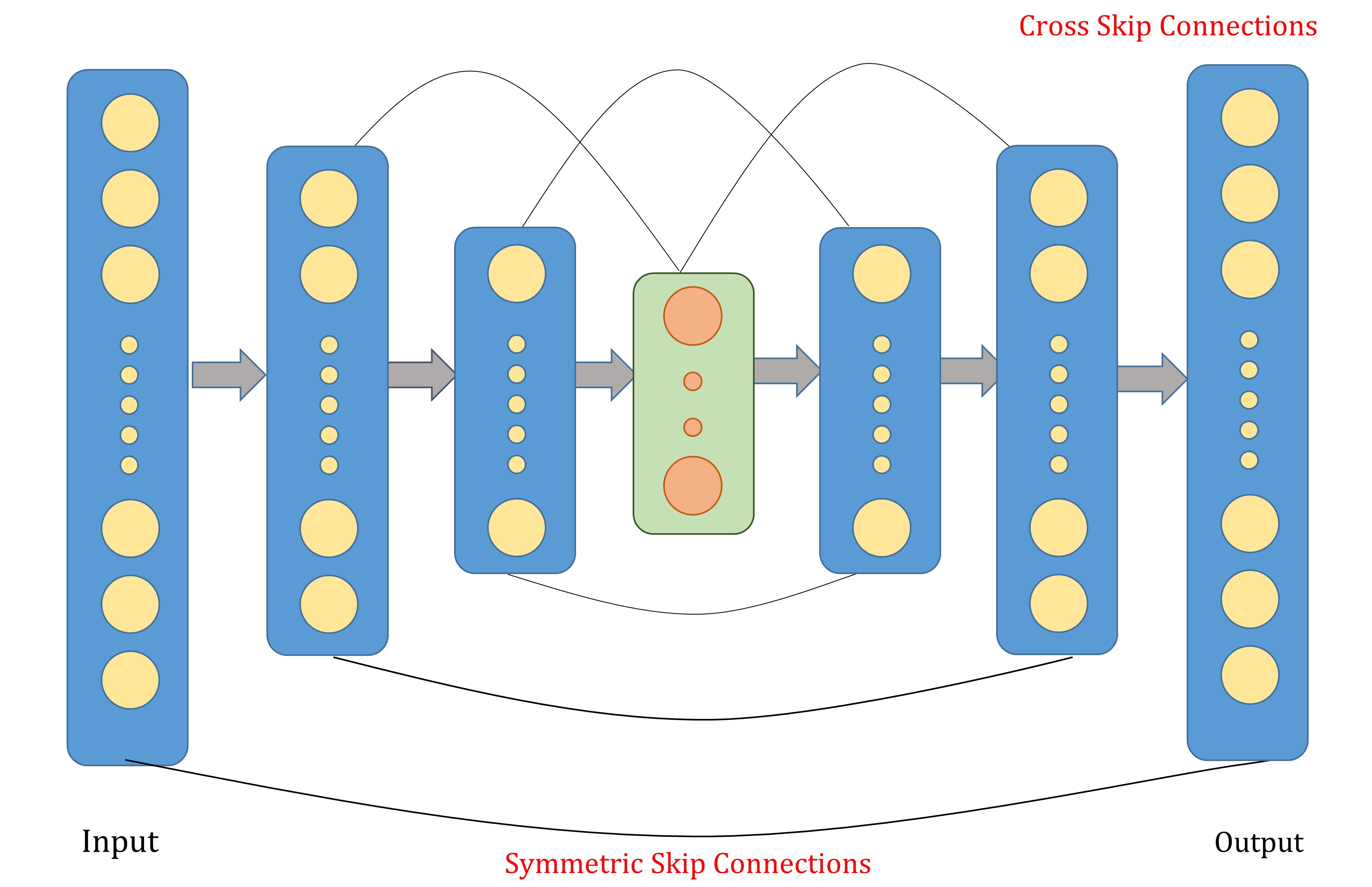}
  \caption{Cross and symmetric shortcut (skip) connections in the proposed residual autoencoder model.}
  \label{fig:skip}
\end{figure}

In the proposed Codean autoencoder model, two types of shortcut (or skip) connections have been introduced in order to create the proposed Residual Codean (R-Codean) Autoencoder. The concept of residual learning has been incorporated in the autoencoder model by including $cross$ and $symmetric$ shortcut connections. As shown in Fig. \ref{fig:skip}, cross shortcut connections are overlapping connections which are made between the alternate layers of the network. Such connections help in preventing the input degradation problem in deep networks. Symmetric skip connections are non-overlapping connections created at a larger distance between the encoder and decoder of the same autoencoder. Symmetric shortcut connections help in passing the image details forward, thereby improving the reconstruction process of the autoencoder. Both these connections help in improving the gradient flow which further enables the model to converge to the optimal parameters. Incorporating these shortcut connections in the Codean autoencoder model described above results in the proposed R-Codean Autoencoder for learning robust feature representations. 


\begin{figure*}[!ht]
  \includegraphics[width=\linewidth]{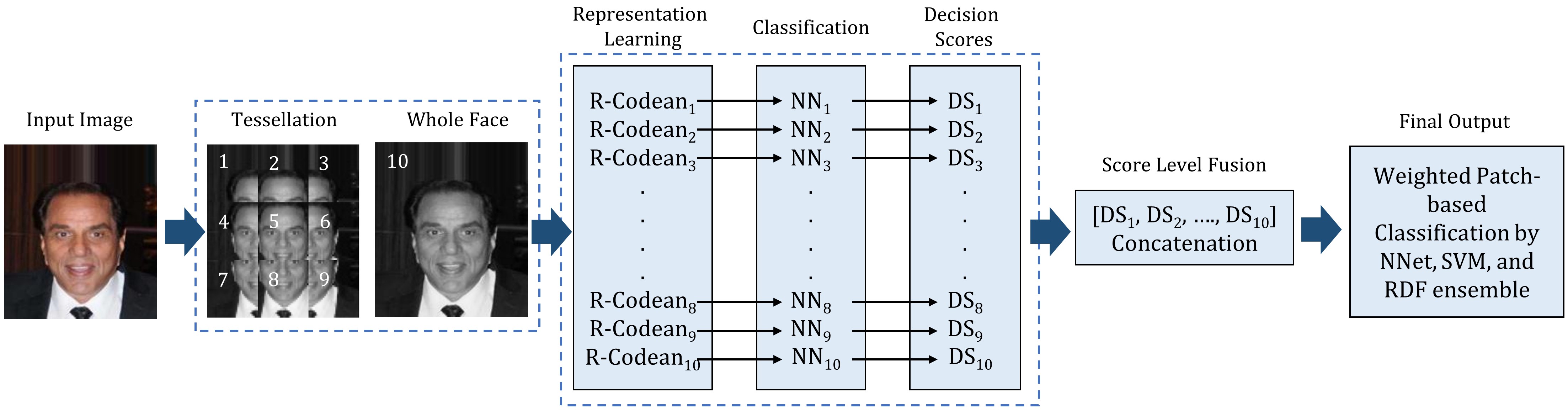}
  \caption{Illustrating the steps involved in the proposed attribute classification pipeline built upon the proposed Residual Cosine Euclidean (R-Codean) Autoencoder.}
  \label{fig:block}
\end{figure*}

\section{Proposed Facial Attribute Prediction Framework using Residual Codean Autoencoder}
The proposed R-Codean autoencoder is utilized for facial attribute prediction. Fig. \ref{fig:block} presents the block diagram of the proposed facial attribute prediction framework. The entire pipeline can broadly be divided into the following components: (i) pre-processing, (ii) feature extraction using R-Codean autoencoder, and (iii) classification. Each step of the pipeline is explained in detail in the following subsections. 

\subsection{Pre-processing of Images}
For a given input sample, the image is geometrically normalized and loose cropped to obtain the face region. The cropped image is down-sampled to $64 \times 64$ and converted to grayscale. In literature, it has often been observed that facial features are encoded both locally, and globally while performing face recognition \citep{patch1, patch2}. Inspired by these findings, a similar approach is followed for encoding facial attributes by tessellating the input image into nine equal overlapping patches (as shown in Fig. \ref{fig:block}). The individual patches as well as the entire image are then used for feature extraction. 

\subsection{Feature Extraction via Proposed R-Codean Autoencoder}
The proposed R-Codean autoencoder is used for the task of feature extraction. One R-Codean autoencoder model is trained for each face image patch, and one for the entire image, thereby resulting in ten independent models (nine patches + 1 full face). Patch based feature learning enables the model to learn specific characteristic of a particular facial region. These local features can then be utilized for predicting face component-specific attributes. Coupled with the learned representation over the entire face image, the ten models provide a holistic (global) as well as local representation of the input sample.

\begin{figure}
\centering
\includegraphics[width=3in]{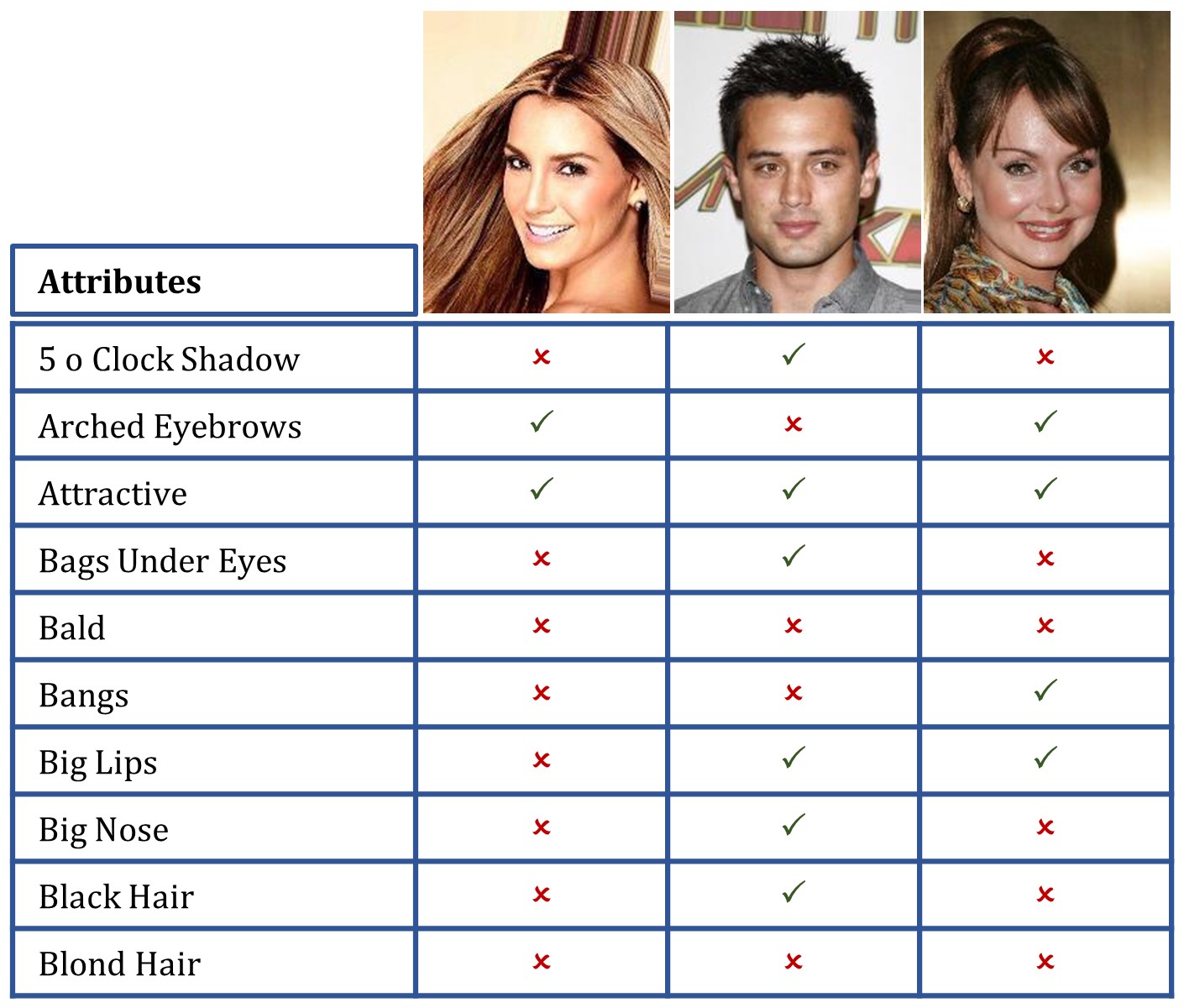}
\caption{Sample images from the CelebA dataset, along with a few attributes. Each attribute is a binary attribute, having value either `1' or `0'. }
\label{fig:celebA}
\end{figure}

\subsection{Weighted Patch-based Classification}
A two-step classification approach is followed in the proposed facial attribute prediction framework. In the first stage, for each R-Codean autoencoder trained on each patch location in the previous step, a two layer neural network classifier is learned using $sigmoid$ activation function in the output layer. Therefore, ten neural networks are trained, one corresponding to each R-Codean autoencoder, i.e., one for each facial patch (9) and one for the full face. $k$ probability scores are obtained from the output layer of each neural network, where $k$ corresponds to the number of facial attributes in the dataset. In the second stage, score fusion is performed by concatenating the scores obtained from the ten neural networks to create a feature vector of length $10 \times k$. This feature vector is then provided as input to an ensemble of Neural Network, Random Decision Forest, and Support Vector Machine for obtaining the final attribute classification. In order to emulate the human tendency of focusing on specific regions for certain attributes, a weighted patch-based mechanism is also incorporated in the classification framework. Patch-based weights are learned for each attribute, such that relevant patches are given higher weights for a given attribute. For instance, in case of \textit{wearing necktie} attribute, patches 7, 8, and 9 (shown in Figure \ref{fig:block}) might contribute higher in the decision making process, as opposed to the other patches. Finally, max-voting is performed on the outputs of the three classifiers in order to obtain the final decision.  

\subsection{Implementation Details}
For all experiments, detected and normalized face images are resized to $64 \times 64$ and converted to grayscale. Images are tessellated in $3\times3$ overlapping patches of dimension $32\times32$. 10 R-Codean autoencoders having three hidden layers with dimensions $[l, l, l]$, are trained on the face patches (and full face). $k$ corresponds to the dimension of the input vector. $\ell_1$-norm regularizer with $\lambda = 0.01$ is incorporated for learning a sparse representation of the data. Each R-Codean autoencoder also incorporates cross shortcut connections between the first and third encoding layers, second encoding and first decoding layer, third encoding and second decoding layer. Symmetric shortcut connections are also introduced between the first encoding and third decoding layer, second encoding and decoding layer, third encoding and first decoding layer. R-Codean is optimized using the adam optimizer \citep{adam} with a decaying learning rate. The initial learning rate was set to 0.001, which decayed by a factor of ten whenever the training loss stagnated. For each R-Codean autoencoder, a two hidden layer neural network of dimension $[\frac{l}{2}, \frac{l}{4}]$ is learned in the first classification stage. The autoencoders and neural networks utilize $ReLU$ activation function in their layers. The autoencoders are implemented in Python based Keras framework \citep{keras} and the classifiers in Scikit-learn library \citep{scikit}. The source code will be made publicly available for the research community.  

\renewcommand{\arraystretch}{1.2}
\begin{table}[t]
\centering
\caption{Comparison with state-of-the-art and existing methods on CelebA and LFWA datasets. Accuracy corresponds to the mean accuracy obtained over all the attributes.}
\label{tab:results}
\begin{tabular}{|p{5cm}|p{1.2cm}|p{1.2cm}|}
\hline
\textbf{Architecture} & \textbf{CelebA} & \textbf{LFWA}         \\     \hline
\hline
Stacked Autoencoder + NNET & 85.60\%  & 76.22\% \\     \hline
3-layer CNN + NNET & 87.39\% & 82.37\%\\ \hline
VGG-Face \citep{parkhi2015deep} + NNET & 85.30\% & 81.04\% \\  \hline
Fine-tuned VGG-Face \citep{parkhi2015deep} + NNET & 87.45\%  & 80.14\% \\  \hline
ResNet \citep{he2016deep} + NNET & 84.03\% & 76.94\%  \\ \hline
Fine-tuned ResNet \citep{he2016deep} + NNET & 86.23\% & 78.98\% \\ \hline
\cite{zhong2016face} & 86.80\% & 84.70\% \\     \hline  
\cite{liu2015deep} & 87.00\%  & 84.00\% \\     \hline
\cite{wang2016walk}& 88.00\% & \textbf{87.00\%}  \\     \hline
\cite{zhong2016leveraging} & 89.80\% & \textbf{85.90\%} \\     \hline  
\cite{rozsa2016facial} & \textbf{90.80}\% & - \\     \hline 
\cite{rudd2016moon} & \textbf{90.94}\% & - \\ \hline
\textbf{Proposed} & \textbf{90.14}\%    & \textbf{84.90\%} \\ \hline
\end{tabular}
\end{table}


\begin{table*}
\begin{center}
\caption{Attribute classification accuracy (\%) of all forty attributes of the CelebA dataset using the proposed facial attribute prediction framework.}
\label{tab:combine}
\begin{tabular}{|l|r||l|r||l|r|}
\hline
\textbf{Attribute} & \textbf{Accuracy} & \textbf{Attribute} & \textbf{Accuracy} & \textbf{Attribute} & \textbf{Accuracy}  \\ 
                 \hline
                 \hline
                 
5'o' Clock Shadow & 92.88 & Arched Eyebrows  & 81.63 &  Attractive  & 79.67 \\
\hline
Bags Under Eyes  & 83.15 & Bald             & 99.52  & Bangs            & 94.51 \\
\hline
Big Lips         & 79.89  & Big Nose         & 83.67 & Black Hair       & 84.80  \\
\hline
 Blond Hair       & 94.97  & Blurry           & 96.57  & Brown Hair       & 82.97 \\
\hline
Bushy Eyebrows   & 91.36  & Chubby           & 95.51  & Double Chin      & 96.45  \\
\hline
 Eyeglasses       & 98.18 & Goatee           & 96.77 & Gray Hair        & 97.92  \\
 \hline
 Heavy Makeup     & 89.72  & High Cheekbones  & 86.74 & Male              & 95.87 \\
 \hline
 Mouth S. Open     & 89.81 & Mustache          & 96.31  & Narrow Eyes  & 90.64  \\ 
\hline
No Beard          & 94.64 & Oval Face         & 76.55 & Pale Skin         & 96.93  \\
\hline
Pointy Nose       & 76.95 & Receding Hairline & 93.64 &  Rosy Cheeks       & 95.32 \\ 
\hline
Sideburns         & 97.60 & Smiling           & 92.83 & Straight Hair     & 81.19 \\
\hline
 Wavy Hair         & 75.42 & Wearing Earrings  & 82.65  & Wearing Hat       & 97.93 \\
\hline
Wearing Lipstick  & 91.96 & Wearing Necklace  & 89.82 & Wearing Necktie   & 95.88 \\
\hline
Young             & 86.63 & - & - & - & - \\  
\hline
\end{tabular}
\end{center}
\end{table*}

\begin{figure*} [h]
\centering
  \includegraphics[clip, trim = 0.1cm 0.1cm 0.1cm 0.1cm, width=7.4in]{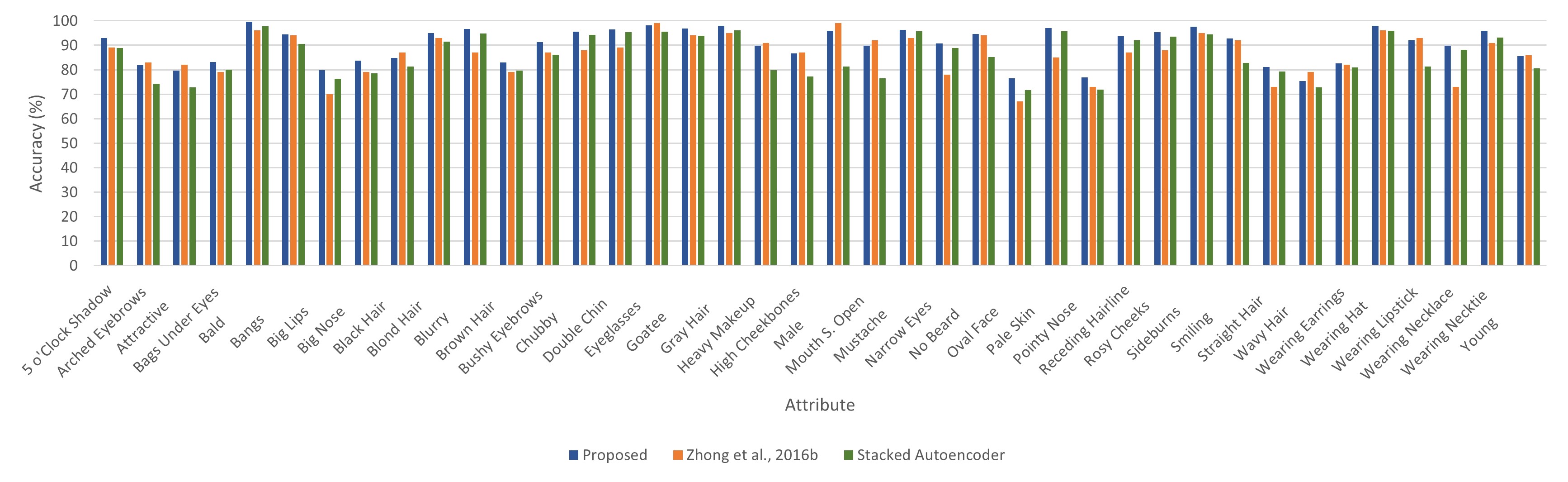}
  \caption{Bar graph illustrating the performance of the proposed framework in comparison with current state-of-the-art technique \citep{zhong2016leveraging} and stacked autoencoder. Accuracies have been reported for all $40$ attributes for the CelebA dataset.}
  \label{fig:bar}
\end{figure*}

\section{Experimental Results and Analysis}
\label{sec:res}
The proposed approach is evaluated on large-scale Celeb Faces Attributes (CelebA) and Labeled Faces in the Wild Attributes (LFWA) datasets \citep{liu2015deep}. The CelebA dataset contains 10,000 identities, each of which have twenty images. Therefore, the dataset contains a total of 2,00,000 images. LFWA contains 13,233 images pertaining to 5,749 subjects. Each image in both the datasets is annotated with forty binary face attributes such as Male, Young, Bangs, Gray Hair, and five facial landmark key points. Fig. \ref{fig:celebA} presents some sample images of CelebA dataset. 

The efficacy of the proposed model has been evaluated on existing commonly used experimental protocols \citep{liu2015deep, lianzhi2016deep, gunther2016affact, zhong2016leveraging}. For the CelebA dataset, the entire dataset is partitioned into three parts, where the first 1,60,000 images are used to train the autoencoders, and next 20,000 images are used to train the ensemble classifier. The images of the remaining 1,000 identities (with 20,000 images) are used for testing. On the other hand, LFWA dataset is divided into two, one partition is used for training, while the second forms the test set.

\subsection{Comparison with Other Deep Learning Models} 
The proposed facial attribute framework, built upon the proposed R-Codean autoencoder yields an overall mean classification accuracy of $\mathbf{90.14\%}$ and \textbf{84.90\%} over the 40 attributes of CelebA and LFWA datasets, respectively. Table \ref{tab:results} presents the mean accuracy of the proposed framework, along with other comparative architectures. In order to compare the performance of other deep model architectures, experiments are performed using a Stacked Autoencoder, Convolutional Neural Network (CNN), and existing state-of-the-art CNN models of ResNet \citep{he2016deep} and VGG-Face \citep{parkhi2015deep}, while using a neural network for classification. For VGG-Face and ResNet, comparison has been performed with both pre-trained models, as well as fine-tuned models. Fine-tuning is performed on the pre-trained models using the training partition of each dataset, while the Stacked Autoencoder model is trained from scratch with the available training data. It can be observed that the proposed R-Codean autoencoder based pipeline achieves an improvement of more than 4\% on the CelebA and 7\% on the LFWA database, as compared to stacked autoencoders. The proposed pipeline also presents improved results as compared to several existing state-of-the-art CNN models (with and without fine-tuning). Specifically, an improvement of more than 4\% is observed in comparison to the ResNet model, along with similar results for VGG-Face features.

\begin{figure*} [t]
\centering
  \includegraphics[width=7.2in]{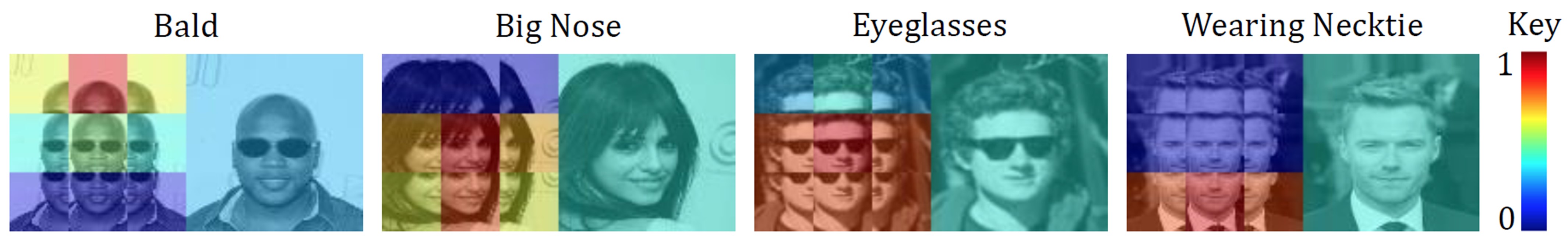}
  \caption{Samples illustrating the effect of weighted patch-based attribute classification. For certain attributes, specific regions are more focused and hence more weights are assigned (learned). For example, in case of \textit{bald}, high weight is associated with patches containing the head region of the face image, while low weight is given to other parts of the image.  }
  \label{fig:patch}
\end{figure*}

\subsection{Comparison with State-of-the-Art Techniques} 
Comparison has also been performed with existing state-of-the-art architectures present in the literature. Table \ref{tab:results} presents the accuracies of recently proposed architectures, taken directly from their publications. It can be observed that the proposed R-Codean autoencoder based approach achieves a comparable mean classification accuracy with respect to the current state-of-the-art approach \citep{rudd2016moon} on the CelebA dataset. It is important to note that while the existing architectures incorporate Convolutional Neural Networks in their pipeline, this is the first work achieving comparable classification performance using autoencoders. Training R-Codean autoencoder requires only $20$ seconds per epoch on a workstation powered with NVIDIA K20 GPU and $64$GB RAM. Moreover, an unseen input sample takes less than a second for the entire attribute prediction pipeline. Table \ref{tab:combine} provides the individual accuracy of each attribute obtained on the CelebA dataset. Fig. \ref{fig:bar} presents the bar graph, depicting comparison with \cite{zhong2016leveraging} and stacked autoencoder features over all 40 attributes. Since the attribute-specific accuracy is not provided by \cite{rozsa2016facial} or \cite{rudd2016moon}, attribute wise comparison is not possible. 

\begin{table} [h]
\caption{Evaluation of the proposed R-Codean based framework for performing facial attribute classification on CelebA dataset.}
\label{tab:compare}
\begin{tabular}{|p{6cm}|c|}
\hline
\textbf{Architecture} & \textbf{Accuracy}          \\     \hline \hline
\multicolumn{2}{|c|}{\textbf{Effect of Pre-Processing}} \\
\hline
\hline
Learning a Full Face Model Only & 86.86\% \\
\hline
Learning Patch-based Models Only & 88.16\%\\
\hline
\multicolumn{2}{|c|}{\textbf{Effect of Shortcut Connections}} \\
\hline
\hline
With Symmetric Connection only & 87.90\% \\
\hline
With Cross Connection Only &  88.81\%\\
\hline
With No Shortcut Connections & 87.60\%\\
\hline
\multicolumn{2}{|c|}{\textbf{Effect of Loss Function}} \\
\hline
\hline
With Euclidean Distance (MSE) only & 88.30\%\\
\hline
With Cosine Distance only & 85.10\%\\
\hline
\multicolumn{2}{|c|}{\textbf{Effect of Patch-based Weighting}} \\
\hline
\hline
Without Patch-based Weighting & 89.42\%\\
\hline
\multicolumn{2}{|c|}{\textbf{Effect of Classification Ensemble}} \\
\hline
\hline
Support Vector Machine Only & 88.81\%\\
\hline
Neural Network Only & 88.30\%\\
\hline
Random Decision Forest Only & 88.84\%\\
\hline
\hline
\textbf{Proposed Framework with R-Codean} & \textbf{90.14\%}\\
\hline
\end{tabular}
\end{table}

\subsection{Analysis of the Proposed Facial Attribute Prediction Pipeline} 
In order to thoroughly evaluate the proposed R-Codean autoencoder based pipeline for facial attribute prediction, experiments have been performed on the CelebA dataset to evaluate each component of the same. Table \ref{tab:compare} presents the mean classification accuracies for different variations of the proposed framework. To re-iterate, in the pre-processing component face tessellation is performed, and models are trained for both full face, as well as independent patches, in order to encode both local as well global features. The framework is analyzed by learning features on only the full face \textit{or} the tessellated patches, instead of both. As can be observed from Table \ref{tab:compare}, both the techniques independently yield lower results as compared to their combination, thereby strengthening our hypothesis of utilizing both local and global features for facial attribute prediction.  

Further evaluations are performed on the R-Codean autoencoder model by understanding the effect of shortcut connections, and the combined loss function. As can be seen from Table \ref{tab:compare}, removing residual shortcut connections from the model results in a drop of around 3\%. A similar drop in accuracy can be observed upon incorporating only a single kind of shortcut connection as opposed to both. In order to evaluate the efficacy of the loss function of R-Codean autoencoder, where the loss function is a combination of Euclidean distance (MSE), as well as the Cosine distance, comparison has been performed with models having only Cosine or Euclidean distance based loss function as well. An increase of at most 5\% is observed upon incorporating the magnitude (Euclidean distance) as well as the direction (Cosine distance) in the feature learning process. At the classification level, the proposed framework utilizes an ensemble of Neural Network, Support Vector Machine, and Random Decision Forest. It can be observed from Table \ref{tab:compare} that upon utilizing each of these classifiers independently, the framework does not yield optimal results. An improvement of close to 1.5\% is observed upon using the proposed ensemble. Comparison has also been performed to understand the effect of training data size on the proposed framework. As can be observed from Fig. \ref{fig:bar}, even when only 50\% of the entire training set is used, the proposed framework yields a classification accuracy of 87.83\%, showcasing a drop of only 2.3\%. This motivates the utility of the proposed R-Codean based framework for less training data as well.  

\begin{figure} [t]
\centering
\includegraphics[width=3.4in]{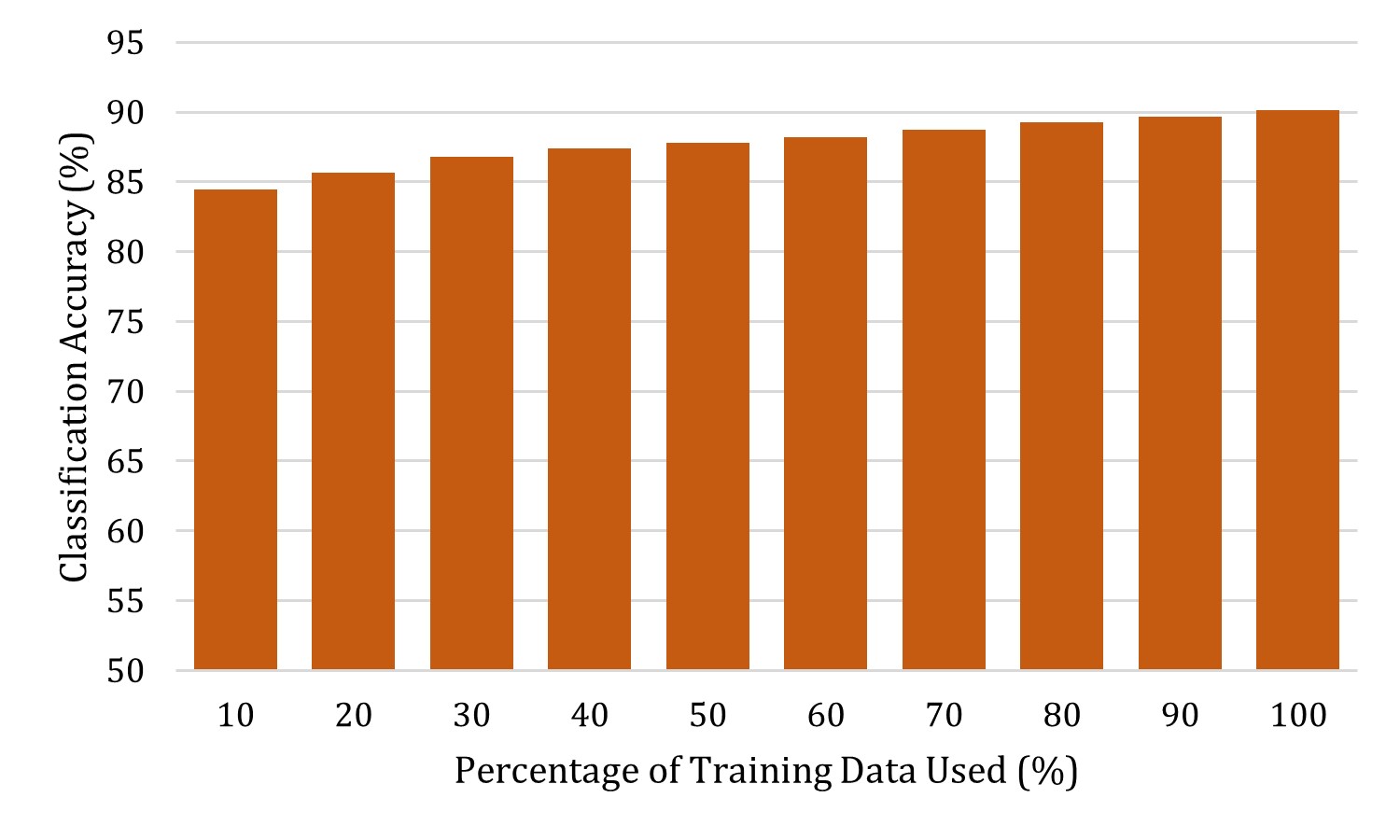}
\vspace{-10pt}
\caption{Bar graph representing the accuracy of the proposed framework with varying percentage of training data, for the CelebA dataset. }
\label{fig:bar}
\end{figure}

To show the effect of learning weights in weighted patch-based classification, Fig. \ref{fig:patch} illustrates some sample attributes, along with the associated weights. The values of weights lie between $0 - 1$ and the illustration is color coded, \textit{red} corresponds to $1$ and \textit{blue} corresponds to $0$. It can be observed that the learned weights are intuitive and the patches which relevant to the given attribute are selected. For instance, in case of \textit{Wearing Necktie}, high weight is given to patches which contain the lower portion of the image. The opposite of this is observed for \textit{Bald}, where high weight is observed for patches which contain the head, while very low weight is associated with the lower portion of the image. 

\begin{figure}
\centering
\subfloat[False Positives]{\includegraphics[height=2in]{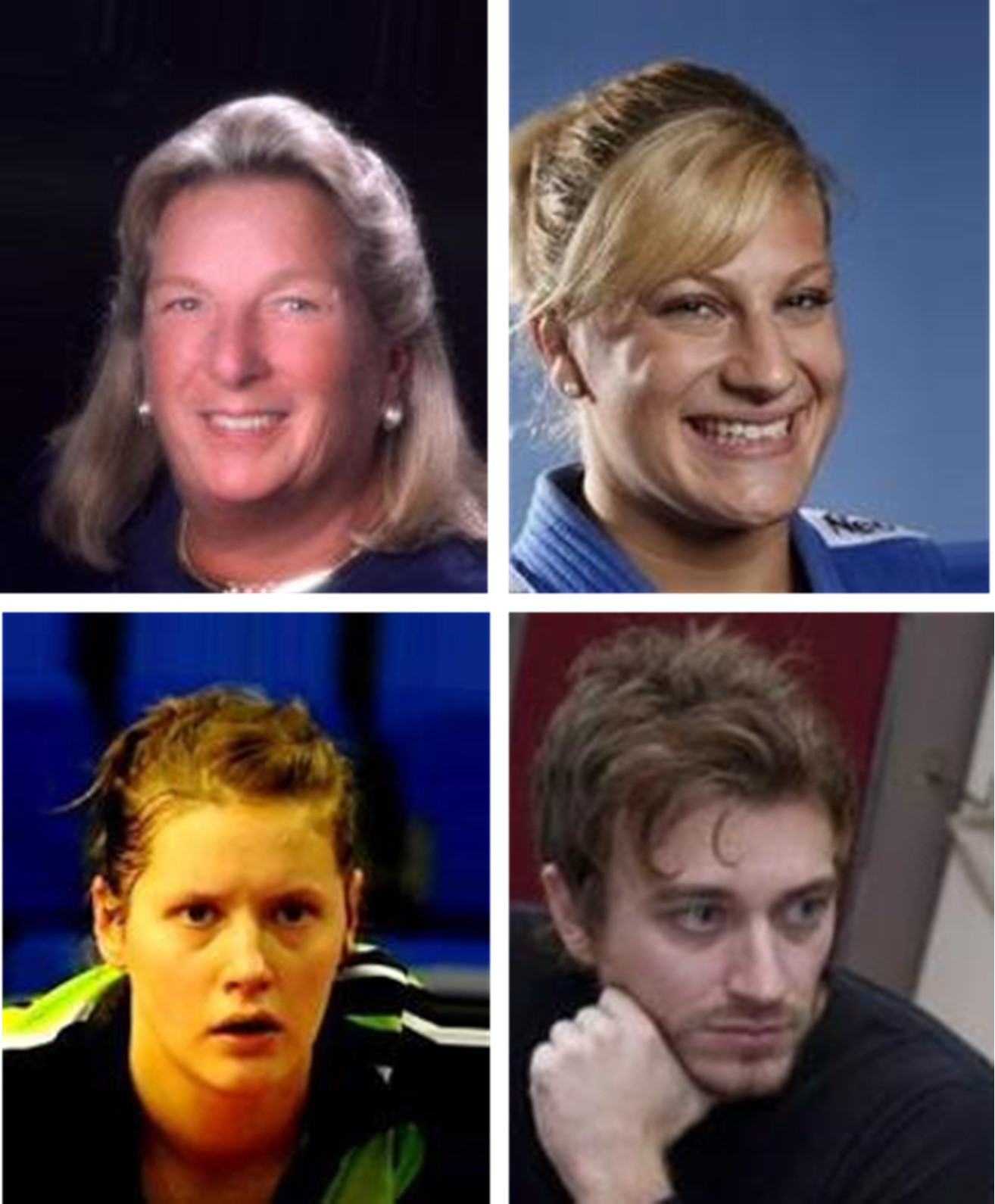}}
\hspace{2pt}
\subfloat[False Negatives]{\includegraphics[height=2in]{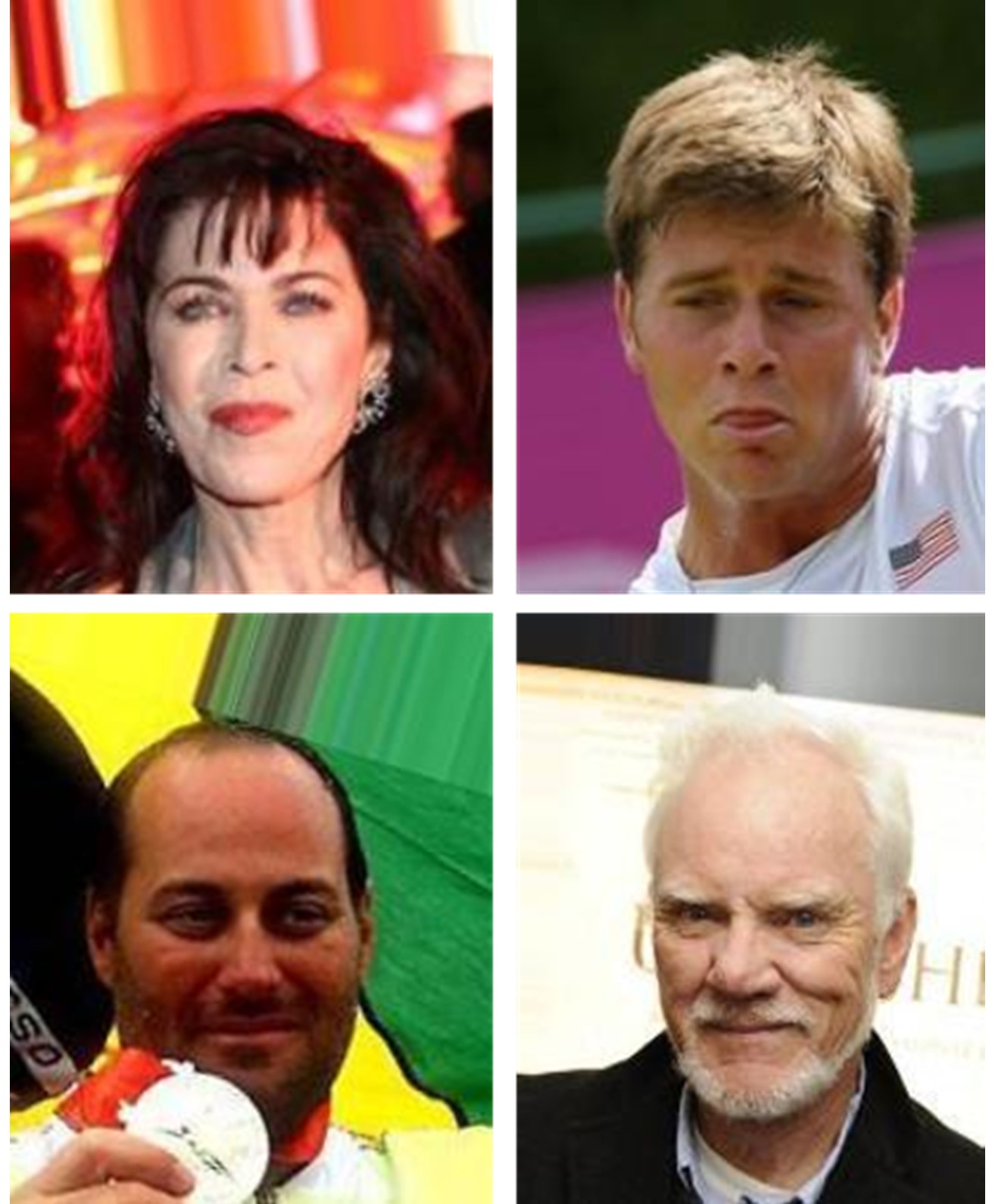}}
\vspace{-5pt}
\caption{Sample mis-classifications of the proposed facial attribute prediction pipeline. (a) corresponds to the false positives, where the images in the first row are falsely predicted as attribute \textit{male} and second row is predicted to have the attribute \textit{wavy hair}. (b) refers to sample false negatives, where the first row corresponds to the attribute \textit{attractive}, and second row corresponds to the attribute \textit{receding hairline}. }
\label{fig:mis}
\end{figure}

Fig. \ref{fig:mis} shows some sample images misclassified by the proposed facial attribute prediction pipeline. Fig. \ref{fig:mis}(a) refers to some sample false positives. These example show that large variations in facial features make the task of facial attribute prediction challenging. Variations in hairstyles across males and females also lead to mis-classifications for attributes such as \textit{male} and \textit{wavy hair} (Fig. \ref{fig:mis} (a)). Fig. \ref{fig:mis}(b) also presents some false negative samples of the proposed framework. It can be observed that subjective attributes such as \textit{attractive} and \textit{receding hairline} are difficult to encode in the trained model. 



\section{Conclusion}
This research aims to address the problem of facial attribute prediction in the wild. A novel formulation for Residual Cosine Euclidean Autoencoder, termed as R-Codean has been presented for the same. The loss function of the proposed R-Codean model consists of a Euclidean distance based term, and a Cosine similarity based term. The model aims to learn representations of the input data, such that the error in direction and magnitude of the input and reconstructed sample is minimized. The Euclidean distance based component of the loss function handles slight variations across pose more efficiently, while the Cosine distance based component models the illumination variations better. Shortcut connections in the R-Codean further ensure that optimal parameters are learned during the feature learning process. An entire framework has been presented for performing facial attribute analysis, which encodes local as well as global facial representations during feature learning process, and also incorporates weighted patch-based classification of attributes. Each component of the framework has been analyzed in order to create the optimal framework for facial attribute prediction. Experimental results on the CelebA and LFWA datasets, and comparison with existing state-of-the-art techniques demonstrate the efficacy of the proposed model. 

\section{Acknowledgment}
The authors thank the reviewers and editors for their valuable comments and feedback. This research is partially supported by MEITY (Government of India), India. M. Vatsa and R. Singh are partially supported through Infosys Center for Artificial Intelligence. 

\bibliographystyle{model2-names}
\bibliography{refs}

\end{document}